\documentclass[10pt,twocolumn,letterpaper]{article}

\usepackage{3dv}
\usepackage{times}
\usepackage{epsfig}
\usepackage{graphicx}
\usepackage{amsmath}
\usepackage{amssymb}

\usepackage{subfig}
\usepackage{balance}
\graphicspath{{figures/}}\usepackage{float}
\usepackage{multirow}
\usepackage[table,xcdraw]{xcolor}

\newcommand{\struct}{\mathbf{S}}
\newcommand{\inputpose}{\mathbf{W}}
\newcommand{\cam}{\mathbf{R}}
\newcommand{\dict}{\mathbf{D}}
\newcommand{\constr}{\mathcal{C}}
\newcommand{\sparsevec}{\boldsymbol{\psi}}
\newcommand{\sparseBlock}{\boldsymbol{\Psi}}
\newcommand{\realSet}{\mathbb{R}}

\usepackage[pagebackref=true,breaklinks=true,colorlinks,bookmarks=false]{hyperref}

\threedvfinalcopy 

\def\threedvPaperID{190} 

\ifthreedvfinal\fi
\begin{document}

\title{High Fidelity 3D Reconstructions with Limited Physical Views}


\author{Mosam Dabhi$^{1}$ \hspace{-0.6cm}
\and Chaoyang Wang$^{1}$ \hspace{-0.6cm}
\and Kunal Saluja$^{2}$ \hspace{-0.6cm}
\and L{\'a}szl{\'o} A. Jeni$^{1}$ \hspace{-0.6cm}
\and Ian Fasel$^{2}$  \hspace{-0.6cm}
\and Simon Lucey$^{1,3}$ \\
\vspace{-0.49cm}
\and $^{1}$ Carnegie Mellon University \hspace{-0.6cm}
\and $^{2}$ Apple Inc. \hspace{-0.6cm}
\and $^{3}$ The University of Adelaide  \\
\vspace{-0.49cm}
\and {\fontfamily{qcr}\selectfont \textcolor{magenta}{https://sites.google.com/view/high-fidelity-3d-neural-prior}
 
}
}

\maketitle


\begin{abstract}
Multi-view triangulation is the gold standard for 3D reconstruction from 2D correspondences given known calibration and sufficient views. However in practice, expensive multi-view setups -- involving tens sometimes hundreds of cameras -- are required in order to obtain the high fidelity 3D reconstructions necessary for many modern applications. In this paper we present a novel approach that leverages recent advances in 2D-3D lifting using neural shape priors while also enforcing multi-view equivariance. We show how our method can achieve comparable fidelity to expensive calibrated multi-view rigs using a limited (2-3) number of uncalibrated camera views. 
\end{abstract}

\vspace{-0.6cm}
\section{Introduction} \label{sec: intro}
\vspace{-0.14cm}
Triangulation refers to determining the location of a point in 3D space from projected 2D correspondences across multiple views. In theory, only \textit{two} calibrated camera views should be necessary to accurately reconstruct the 3D position of a point. However, in practice, the effectiveness of triangulation is heavily dependent upon the accuracy of the measured 2D correspondences, baseline, and occlusions. As a result expensive and cumbersome multi-view rigs, sometimes involving hundreds of cameras and specialized hardware, are currently the method of choice to obtain high fidelity 3D reconstructions of non-rigid objects~\cite{panoptic}.

Deep learning has provided an alternate low-cost strategy by posing the 3D reconstruction problem as a supervised 2D-3D lifting problem -- allowing for effective reconstructions with as little as a single view. Recently, there have been several breakthrough works -- notably Deep NRSfM~\cite{deep_nrsfm_pp} and C3DPO~\cite{c3dpo} -- allowing this problem to be treated in an unsupervised manner that requires ONLY 2D correspondences (i.e. no 3D supervision) greatly expanding the utility and generality of the approach. These unsupervised methods make up for the lack of physical views by instead leveraging large offline datasets containing 2D correspondences of the object category of interest. Unlike classical triangulation, these correspondences do not need to be rigid or even stem from the same object instance. Although achieving remarkable results, these deep learning methods to date have not been able to compete with the fidelity and accuracy of multi-view rigs that employ triangulation.

\begin{figure*}[!t]
	\centering
	\includegraphics[width=0.9\linewidth]{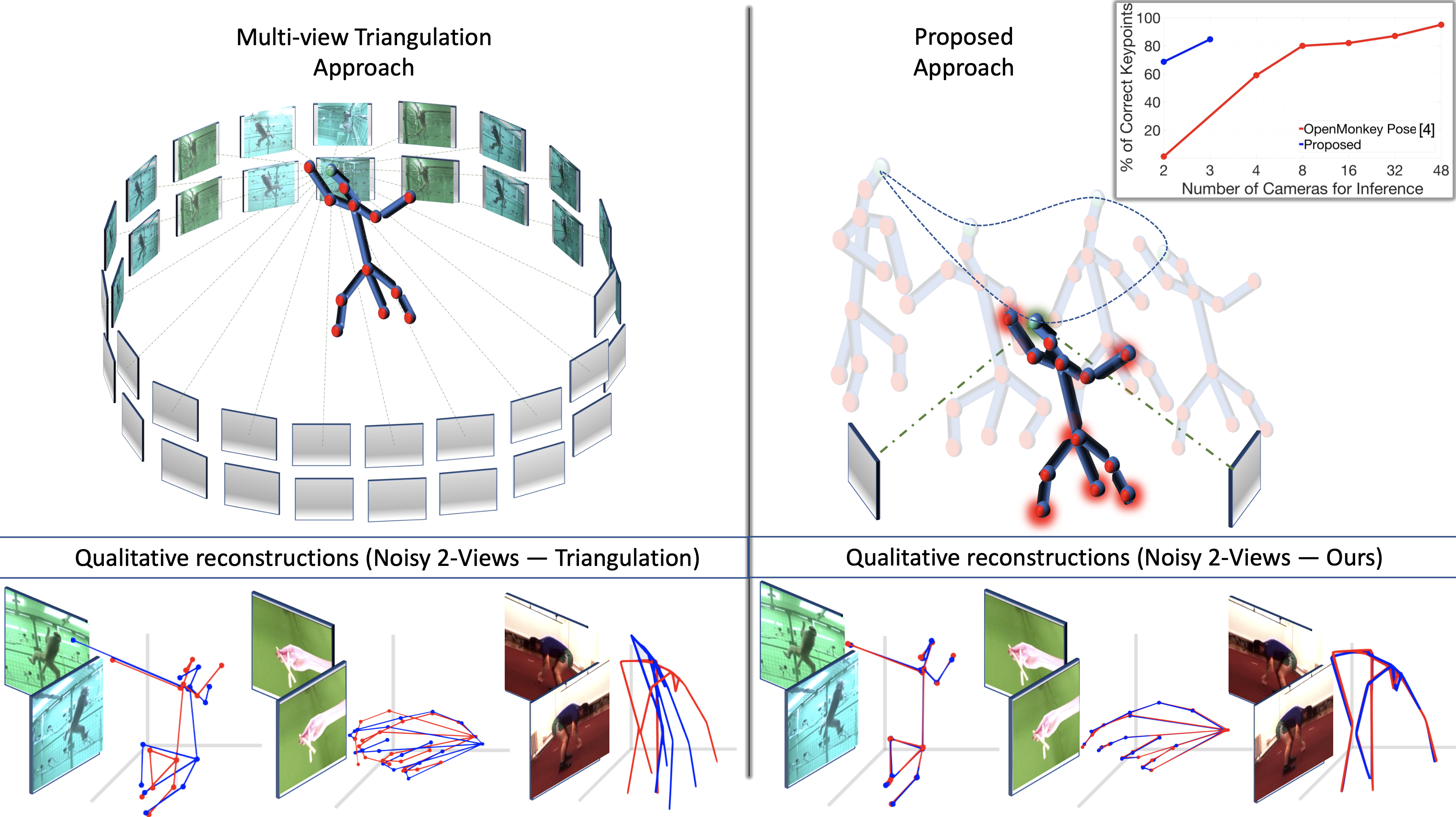}
	\caption{A traditional multi-view setup relies on the concept of triangulation with the assumption that the point being reconstructed is static in time -- requiring a large number of physical views (i.e. cameras) to ensure a high fidelity reconstruction. Our approach utilizes multi-view 2D correspondences taken at the same (rigid) and different (non-rigid) points in time via a neural shape prior. Empirically (see the plot in top-right), we demonstrate that our approach can achieve comparable fidelity to expensive multi-view rigs using only two physical views. Blue lines depict the triangulation and proposed approaches (left vs.~right, respectively) with as little as two physical views, and red lines show the corresponding 3D ground-truth.}
	\label{fig:overview_final} 
	\vspace{-0.3cm}
\end{figure*}

Although the methods using deep learning for single view 2D-3D lifting are of prominent research interest -- we argue that multi-view consistency is still crucial for generating 3D reconstructions of high fidelity needed for many real-world applications. To this end, we propose a new multi-view NRSfM architecture that incorporates a neural shape prior while enforcing equivariant view consistency. We demonstrate that this framework is competitive with some of the most complicated multi-view capture rigs -- while only requiring a modest number (2-3) of physical camera views. Our effort is the first we are aware of, that utilizes these new advances in neural shape priors for multi-view 3D reconstruction. Figure~\ref{fig:overview_final} presents a graphical depiction of our approach. Extensive evaluations are presented across numerous benchmarks and object categories including the human body, human hands, and monkey body. To clarify further, we are interested in a problem setup with multiple views that capture different instances of a deformable object – we deal with non-sequential (atemporal) data. 

\noindent \textbf{Motivation:}
In many problems, complex multi-camera rigs may be financially, technologically, or simply practically infeasible. Our work in this paper is motivated by the realization that the simplification of multi-view camera rigs -- in terms of (i) the number of physical views, and (ii) the need for calibration -- could open the door to a wide variety of applications including entertainment, neuroscience, psychology, ethology, as well as several fields of medicine~\cite{motivation_1,motivation_2,motivation_3,motivation_4,motivation_5}.

\noindent \textbf{Background:}
One of the most notable multi-view rigs for high-fidelity 3D reconstruction is the PanOptic studio~\cite{panoptic}, which contained $480$ VGA cameras, $31$ HD Cameras, and $10$ RGB+D sensors, distributed over the surface of a geodesic sphere with a $5.49$m diameter. This setup also required specialized hardware for storage and gen-lock camera exposures and was aimed initially at human pose reconstruction. Despite its cost and complexity, the fidelity of the 3D reconstructions from PanOptic studio has motivated similar efforts across industry and academia. Of particular note is a recent effort that employed $62$ hardware synchronized cameras to capture the pose of Rhesus Macaque monkeys~\cite{openmonkey}. Other notable efforts include~\cite{canine_pose} for dogs, ~\cite{human36m} for human body, and~\cite{pie,multipie} for the human face.  

\noindent \textbf{Limitations:}
Classical multi-view triangulation can infer 3D structure solely from the rigid 2D correspondences stemming from the physical cameras at a single time instant. If given sufficient calibrated physical camera views, it remains the gold standard for 3D reconstruction. However, if calibration is unknown or the number of views is sparse, we argue that the proposed approach is of significant benefit. Our approach, though is limited in comparison to triangulation as it requires multi-view 2D correspondences taken at the same (rigid) and different (non-rigid) points in time during the learning/optimization process. 

A strength of the proposed approach, however, is that the non-rigid 2D correspondences are treated atemporally (\textit{i.e.} the temporal ordering of the non-rigid correspondences is completely ignored). This means that once the network weights of our approach have been learned -- high-fidelity 3D estimates of the structure and cameras can be obtained in real-time from the very first frame. Changes in sampling rate or dynamics between training and run-time have no bearing on performance - for the same reason. Further, just like classical multi-view triangulation, our approach requires no 3D supervision and hence relies only on 2D correspondences. The proposed approach also assumes known 2D projected measurements so it does not directly leverage pixel intensities. Therefore, our approach can be integrated with any available 2D landmark image detector such as HR-Net~\cite{hrnet}, Stacked Hourglass Networks~\cite{stacked_hourglass}, Integral Pose Regression~\cite{integral_pose_regression}, and others. Finally, the camera is \textbf{not} assumed to be static making the proposed approach agnostic to camera movements.

\vspace{-0.2cm}
\section{Related Work} \label{sec: rel_work}

\vspace{-0.2cm}
\paragraph{Multi-view approaches:} \label{sec: rel_multiview_modern}
Multi-view triangulation~\cite{hartley_multiview_geometry} has been the method of choice in the context of large-scale complex rigs with multiple cameras~\cite{panoptic,openmonkey,pie,multipie} for obtaining 3D reconstruction from 2D measurements. The number of views, 2D measurement noise, baseline, and occlusions bound the fidelity of these 3D reconstructions. These time-synchronized multiple physical views also come at considerable cost and effort. 
\noindent Recent work by Iskakov et al.~\cite{learnable-triangulation} and others~\cite{remelli2020lightweight,kadkhodamohammadi2021generalizable,tome,Pavlakos} have explored how supervised learning can be used to enhance multi-view reconstruction. Similarly, work by Rhodin et al.~\cite{rhodin-monocular-from-multiview} and Kacobas et al.~\cite{self-supervise-3d} attempted to use supervised and self-supervised learning, respectively, to infer 3D geometry from a single physical camera view. An obvious drawback to these approaches is that one is required to have intimate 3D supervision of the object before deployment -- a limitation that modern multi-view rigs are not faced with. None of these approaches are as general as the one we are proposing. For example, nearly all these prior works deal solely with the reconstruction of the human pose as they are heavily reliant upon peripheral 3D supervision. 

 
\vspace{-0.5cm}
\paragraph{2D to 3D Lifting:}
NRSfM~\cite{low_rank_2} aims to reconstruct the 3D structure of a deforming object from 2D correspondences observed from multiple views. While the object deformation has classically been assumed to occur in time~\cite{akhter2009nonrigid,kumar2020non,park20103d,convolutional_sparse_lucey,kumar2019superpixel,pollefeys1999self}, the vision community has increasingly drawn attention to \emph{atemporal} applications -- commonly known as unsupervised 2D-3D lifting. These temporal approaches rely on the sequential motion of objects, our approach on the other hand is much more unconstrained -- accepting uncalibrated atemporal 2D instances. Advances in unsupervised learning based approaches to 2D-3D lifting~\cite{deep_nrsfm,c3dpo} have seen significant improvements in their robustness and fidelity across a broad set of object categories and scenarios. These recent advances to date have only been applied to problems where there is only a single view (i.e. monocular) of the object at a particular point in time. Our approach is the first -- to our knowledge -- to leverage these advancements for 3D reconstruction when there are multi-view measurements taken at the same instance in time.

\vspace{-0.1cm}
\section{Preliminary}
\vspace{-0.2cm}
\paragraph{Notations} \label{sec: notations}
This paper uses the following notations throughout the manuscript.

\begin{center}
\begin{tabular}{l*{1}{c}r}
	\hline
	Variable type              & Examples \\
	\hline
	Scalar & $s, N, K, L$ \\
	Vector           & $\mathbf{s}, \sparsevec, \boldsymbol{\lambda}$  \\
	Matrix   & $\inputpose, \mathbf{S}, \cam, \mathbf{t}, \dict$ \\
	Function & $\boldsymbol{f_{e}}, \boldsymbol{f_{d}}, \boldsymbol{g}$ \\
	$l^{th}$ layer & $\quad \hspace{-3.5mm} ^{l}\sparsevec, \quad \hspace{-3.5mm} ^{l}\dict$, $\quad \hspace{-3.5mm} ^{l}\boldsymbol{\lambda}$ \\
	$n^{th}$ instance & $\inputpose^{(n)}, \struct^{(n)}, \sparsevec^{(n)}, \boldsymbol{\lambda}^{(n)}$\\
	$k^{th}$ view  & $\inputpose_{k}, \cam_{k}, \sparsevec_{k}$\\
\end{tabular}
\end{center}

Any different signs utilized to explain a mathematical phenomenon other than the ones described above would be explicitly defined wherever deemed necessary.

\begin{figure}[!t]
	\centering
	\includegraphics[width=1\linewidth]{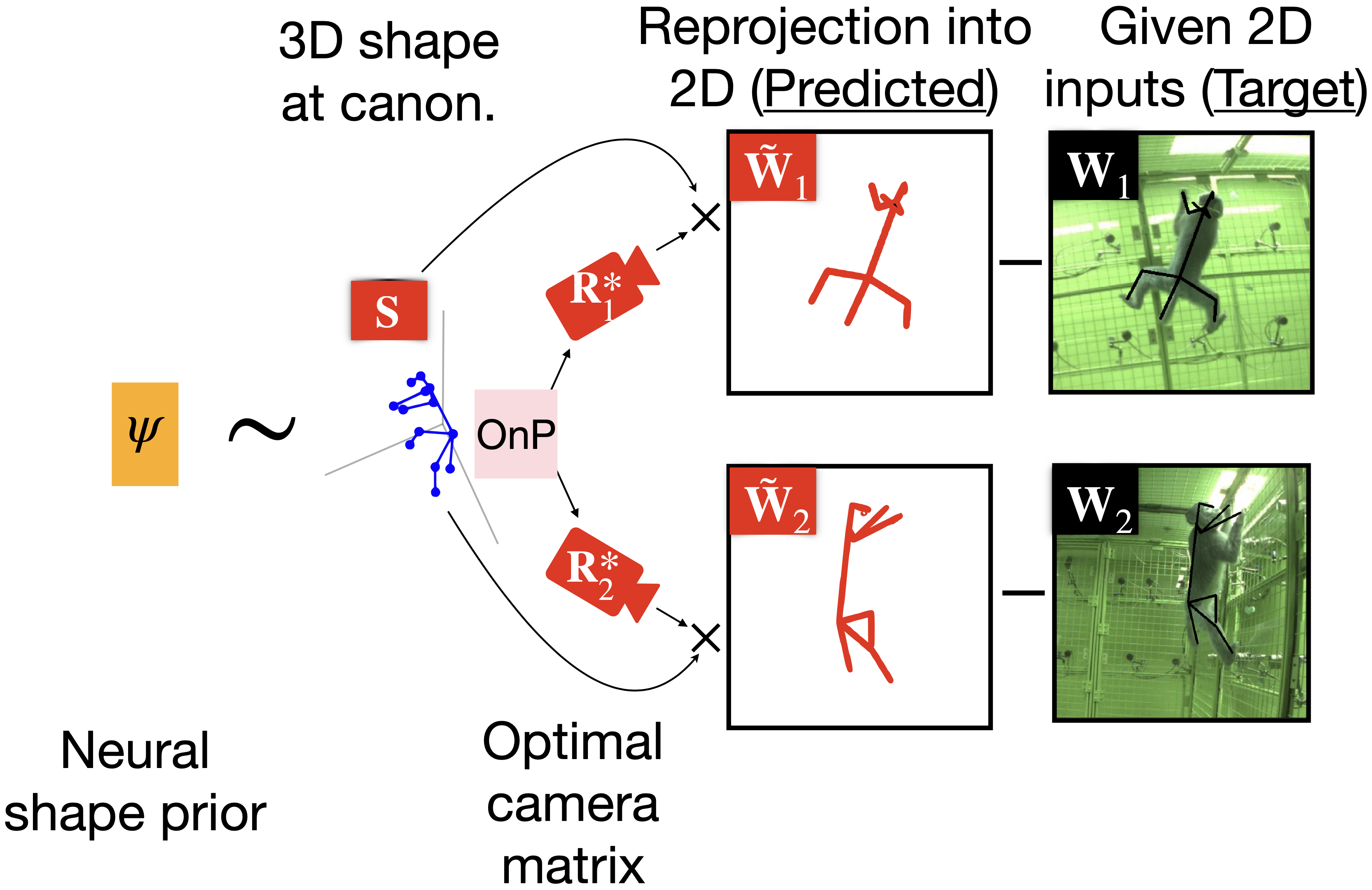}
 	\caption{Two views statistical shape prior. The 3D structure $\struct$ is drawn from a statistical shape distribution using neural shape priors and consequently projected to $2$ views using the cameras $\cam_{k}^{*}$ $\forall k \in [1,2]$ -- calculated through OnP formulation~\cite{onp-formulation}. The proposed approach minimizes the 2D projection error between the predicted 2D projections $\tilde{\inputpose}_{k}$ and target (input) 2D projections $\inputpose_{k}$.
 	\vspace{-1.3em}
}
\label{fig:main_idea}
\end{figure}

\vspace{-1.3em}
\paragraph{Problem setup.}
We are interested in a camera rig setup with $K$ synchronized views capturing $N$ instances (samples) of non-rigid objects from the same category. Specifically, we are given a non-sequential (atemporal) dataset containing $N$ multi-view 2D observations $\{ \mathbf{W}_{1}^{(1)},\ldots,\mathbf{W}_{1}^{(N)}; \cdots; \mathbf{W}_{K}^{(1)},\ldots,\mathbf{W}_{K}^{(N)} \}$, where each $\mathbf{W} \in \mathbb{R}^{P\times 2}$ represents 2D location for $P$ keypoints. We want to reconstruct the 3D shape $\mathbf{S}^{(1)}, \ldots,\mathbf{S}^{(N)}$, where each $\mathbf{S}\in\mathbb{R}^{P\times 3}$ for each of the $N$ instances of the object.

\vspace{-1.3em}
\paragraph{Weak perspective projection.}
We assume weak perspective projections, \textit{i.e.} for a 3D structure $\mathbf{S}$ defined at a canonical frame, its 2D projection is approximated as
\begin{align} \label{eq: bilinear_init}
    \mathbf{W} \approx s\mathbf{S}\mathbf{R}_{xy} + \mathbf{t}_{xy}
\end{align}
where $\mathbf{R}_{xy} \in \mathbb{R}^{3 \times 2}$, $\mathbf{t}_{xy} \in \mathbb{R}^{2}$ are the $x$-$y$ component of a rigid transformation, and $s > 0$ is the scaling factor inversely proportional to the object depth if the true camera model is pin-hole. If all 2D points are visible and centered,
$\mathbf{t}_{xy}$ can be omitted by assuming the origin of the canonical frame is at the center of the object. Due to the bilinear form of~\eqref{eq: bilinear_init}, $s$ is ambiguous and becomes up-to-scale recoverable only when $\mathbf{S}$ is assumed to follow certain prior statistics. We handle scale by approximating with an orthogonal projection and solving an Orthogonal-N-Point (OnP) problem~\cite{onp-formulation} to find the camera pose along with the scale, as discussed in Sec.~\ref{sec: OnP_Para}. 


\vspace{-1.3em}
\paragraph{Statistical shape model.}
We assume a linear model for the 3D shapes $\mathbf{S}$ to be reconstructed, \textit{i.e.} at canonical coordinates, the vectorization of $\mathbf{S}$ in Eq.~\eqref{eq: bilinear_init}, denoted $\mathbf{s} = \text{vec}(\mathbf{S}) \in \mathbb{R}^{3P}$ can be written as 
\begin{align}\label{eq: lin_S}
 \mathbf{s} = \dict \sparsevec   
\end{align} where $\dict \in \realSet^{3P \times B}$ is the shape dictionary with $B$ basis and $\sparsevec \in \realSet^{B}$ is the code vector - taking insight from classical sparse dictionary learning methods. The factorization of $\mathbf{S}$ in Eq.~\eqref{eq: lin_S} is ill-posed by nature; in order to resolve the ambiguities in this factorization, additional priors are necessary to guarantee the uniqueness of the solution. Notable priors include the assumption of $\mathbf{S}$ being $(i)$ low rank~\cite{low_rank_dai_1, low_rank_2,low_rank_3, low_rank_4, low_rank_5}, $(ii)$lying in a union-of-subspaces~\cite{uos_1,uos_2,uos_3} $(iii)$ or compressible~\cite{compressible_2,compressible_3,compressible_1}. The low-rank assumption becomes infeasible when the data exhibits complex shape variations, the Union-of-subspaces NRSfM methods have difficulty clustering shape deformations and estimating affinity matrices effectively just from 2D observations. Finally, the sparsity prior allows more powerful modeling of shape variations with a large number of subspaces but suffers from sensitivity to noise.

\vspace{-1.4em}
\paragraph{Neural Shape Prior}

Our neural shape prior is an approximation to a hierarchical sparsity prior introduced by Kong et al.~\cite{deep_nrsfm}, where each non-rigid shape is represented by a sequence of hierarchical dictionaries and corresponding sparse codes. Other neural shape priors -- such as C3PDO~\cite{c3dpo} -- could be entertained as well but we chose to employ Kong et al.'s method due to its simplicity with respect to enforcing multi-view equivariant constraints. The approach in~\cite{deep_nrsfm} maintains the robustness of sparse code recovery by utilizing overcomplete dictionaries to model highly deformable objects consisting of large-scale shape variation. Moreover, if the subsequent dictionaries in this multi-layered representation are learned properly, they can serve as a filter such that only functional subspaces remain and the redundant are removed. Due to the introduction of multiple levels of dictionaries and codes in the following section, we will abuse the notation of $\dict, \sparsevec$ by adding left superscript $1$, \textit{i.e.} $\quad \hspace{-3.5mm} _{}^{1}\dict, \quad \hspace{-3.5mm} _{}^{1}\sparsevec$ indicating that they form the first level of hierarchy. Assuming the canonical 3D shapes are compressible via multi-layered sparse coding with $l \in L$ layers, the shape code $_{}^{1}\sparsevec$ is constrained as
\vspace{-0.3cm}
\begin{align} 
\begin{split}
    \mathbf{s} &= \quad \hspace{-3.5mm} ^{1}\dict \quad \hspace{-3.5mm} ^{1}\sparsevec  \\
    ^{1}\sparsevec &= \quad \hspace{-3.5mm} ^{2}\dict _{}^{2}\sparsevec \\
    & \vdots \\
    ^{L-1}\sparsevec &= \quad \hspace{-3.5mm} ^{L}\dict ^{L}\sparsevec \\
    \text{s.t. } \text{ } \| ^{l}\sparsevec \|_{1} \leq \quad \hspace{-3.5mm}^{l}\boldsymbol{\lambda} &\text{ , } ^{l}\sparsevec \geq \mathbf{0} \text{ , } \forall l \in \{ 1, \ldots, L \} \label{eq: init_decom}     
\end{split}
\end{align} where $^{l}\dict \in \realSet^{\quad \hspace{-3.5mm} ^{l-1}B \times \quad \hspace{-3.5mm} ^{l}B}$ are the hierarchical dictionaries, $l$ is the index of hierarchy level, and $^{l}\boldsymbol{\lambda}$ is the scalar specifying the amount of sparsity in each level. Thus, the learnable parameters are $\Theta = \{ \cdots, \quad \hspace{-3.5mm} ^{l}\dict, \quad \hspace{-3.5mm} ^{l}\boldsymbol{\lambda}, \cdots \}$.  
The single set of parameters $\Theta$ are fit {\it jointly} along with the sparse codes, rotation matrices, and structures $\mathbf{S}$ for each instance in the dataset. 
Jointly constraining each instance via a common set of weights (the ``neural prior'') makes this work more akin to classic factorization methods, in which both the shared factors and the weightings for each instance are jointly inferred, rather than to network training approaches which aim to find weights that generalize well when later used to perform inference on unseen data.

\vspace{-1.4em}
\paragraph{Factorization-based NRSfM.}
Equivalently, the linear model in Eq.~\eqref{eq: lin_S} could be rewritten as \begin{align*}
 \struct = \dict^{\#} (\sparsevec \otimes \mathbf{I}_{3})   
\end{align*} where $\dict^{\#} \in \realSet^{P \times 3B} $ is a reshape of $\dict$ and $\otimes$ denotes a Kronecker product. Applying the camera matrix $\cam_{xy}$ gives the 2D pose. Thus \vspace{-0.3cm} \begin{align*} 
\struct \cam_{xy}   = \dict^{\#} (\sparsevec \otimes \cam_{xy})
\end{align*} 
Substituting the input 2D pose $\inputpose$ from Eq.~\eqref{eq: bilinear_init}, we have
\begin{align}
\begin{split}
\inputpose &= \dict^{\#}\sparseBlock_{xy} \\
\text{s.t. } \text{ } \sparseBlock_{xy} &= \sparsevec \otimes \cam_{xy} \text{ and } \sparsevec \in \constr \label{eq: 2d_projection_equation}      
\end{split}
\end{align} where $\sparseBlock_{{xy}} \in \realSet^{3B \times 2}$ is the sparse block code denoting the first two columns of $\sparseBlock \in \realSet^{3B \times 3}$; and $\constr$ denotes the neural shape prior constraints applied on the code $\sparsevec$, \textit{e.g.}~hierarchical sparsity~\cite{deep_nrsfm} in our case. Conceptually, $\sparseBlock$ is a matrix with rotations and sparse code built into it. Under the unsupervised settings, $\dict, \sparsevec, \cam, \struct$ are all unknowns and are solved under the simplified assumptions that the input 2D poses are obtained through a weak perspective camera projection. We also analytically compute $\cam^{*}$ as a solution to a Orthographic-n-point (OnP) problem that acts as a supervisory signal to $\cam$. Corresponding proof for $\cam^{*}$ is discussed in the supplementary section.

\vspace{-0.1cm}
\section{Approach}
\vspace{-0.2cm}
\begin{figure}[!t]
	\centering
	\includegraphics[width=1\linewidth]{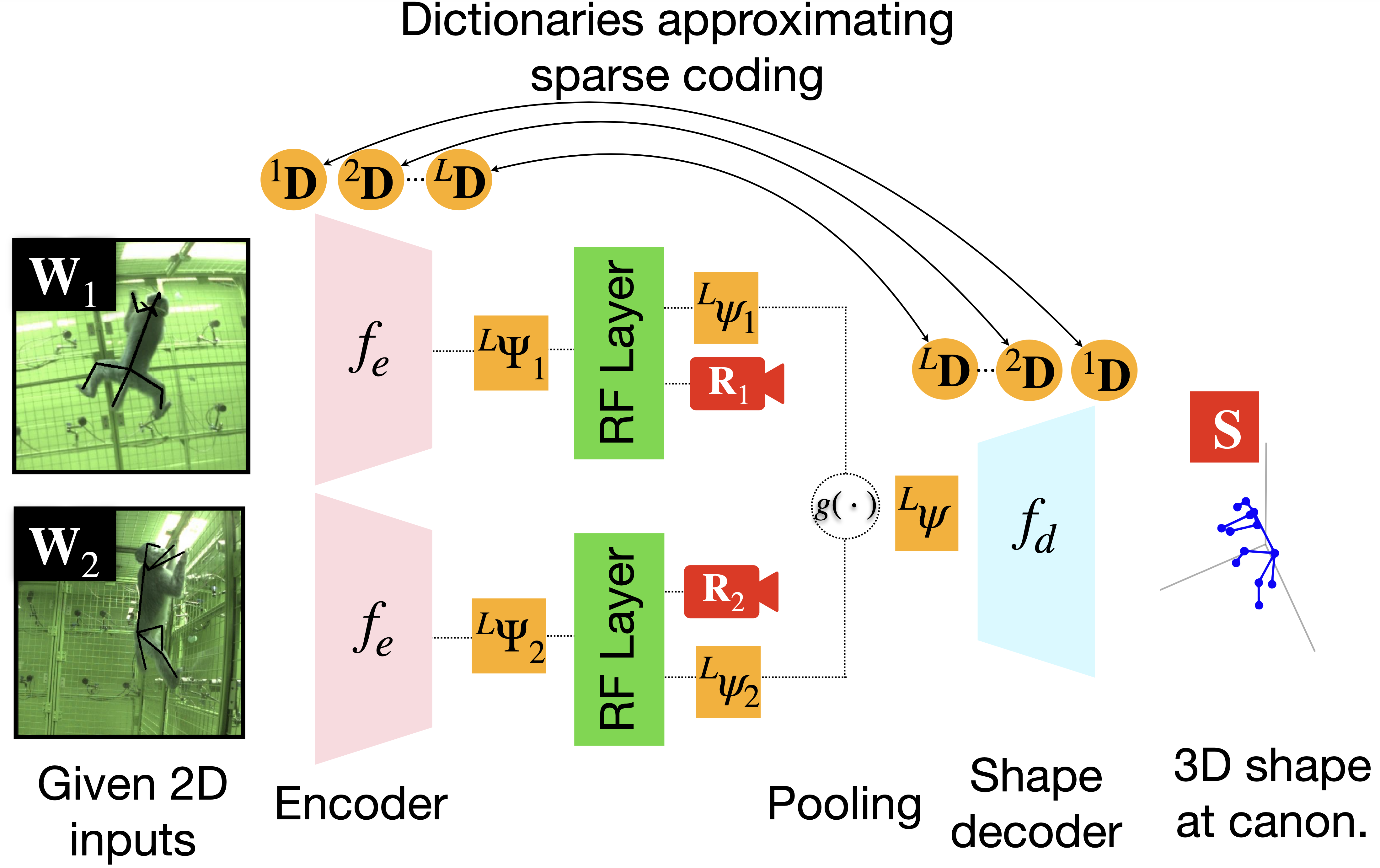}
 	\caption{Architecture showing a $2$-view 3D reconstruction approach (easily extensible to $K>2$ views). The 2D projections from both views $\inputpose_{k}$ $\forall k \in [1,2]$ acts as an input to encoder $\boldsymbol{f}_{e}$ that extracts the block sparse code $\sparseBlock_{k}$ from the corresponding views. A Rotation Factorization (RF) layer at the bottleneck stage shown in green, factorizes the block sparse code into the respective camera matrix $\cam_{k}$ and the unrotated vector sparse code $^{L}\sparsevec_{k}$. The codes are then fused via \textit{pooling} function $\boldsymbol{g}$ into a single code $^{L}\sparsevec$ that acts as an input to the shape decoder $\boldsymbol{f}_{d}$. The shape decoder predicts the 3D structure $\struct$ in the canonical frame while enforcing equivariant view consistency.
 	\vspace{-0.1cm}
 	}
\label{fig:main_arch}
\vspace{-0.4cm}
\end{figure}

\subsection{Bilevel optimization}
\vspace{-0.1cm}
\noindent Given only the input 2D poses in Eq.~\eqref{eq: 2d_projection_equation}, two problems remain to address
\begin{itemize}
    \item How to formulate an optimization strategy to recover $\dict, \sparsevec, \cam, \struct$?
    \vspace{-0.2cm}
    \item How to efficiently pool in $K$ different camera views and enforce equivariance over the predicted $K$ camera matrices and a single 3D structure in canonical frame?
\end{itemize}

We choose to impose neural shape priors through hierarchical sparsity constraints~\cite{deep_nrsfm} literature for approaching a solution to the above problems, with learnable parameters $\Theta$ (see Sec.~\ref{sec: network_approx}). From Eq.~\eqref{eq: 2d_projection_equation}, the learning strategy of multi-view NRSfM problem for $N$ instances with $K$ views is then interpreted as solving the following bilevel optimization problem. Eq.~\eqref{eq: init_decom} leads to relaxation of the following lower-level problem
\begin{align}
\begin{split}
    &\min_{\dict, \Theta} 
    \sum_{k=1}^{K}\sum_{n=1}^{N} \Bigg( \min_{^{l}\sparsevec_{k}^{{(n)}}, \cam_{k}^{(n)}} \| \inputpose_{k}^{(n)} - \quad \hspace{-3.5mm} ^{1}\dict \big(\quad \hspace{-3.5mm} ^{1}\sparseBlock_{k}^{{(n)}} \big) \|_{F} + \\
    &\sum_{l=1}^{L} \quad \hspace{-3.5mm} ^{l}\boldsymbol{\lambda} \| ^{l}\sparseBlock_{k}^{{(n)}} \|_{F}^{(3 \times 2)}
    + \sum_{l=2}^{L} \| (^{l-1}\sparseBlock_{k}^{{(n)}}) - \quad \hspace{-3.5mm} ^{l}\dict \quad \hspace{-3.5mm} (^{l}\sparseBlock_{k}^{{(n)}} ) \|_{F} \Bigg) \label{eq: lower_level_problem}    
\end{split}
\end{align}

\noindent where the first expression in~\eqref{eq: lower_level_problem} minimizes the 2D projection error, the second expression enforces sparsity, and the third expression fits each dictionary in the hierarchy to the dictionary representation in the preceding layer. Minimizing the block Frobenius norm of $\sparseBlock$ is equivalent to minimizing the $L_1$ norm of the vector sparse code $\sparsevec$ because $\| \sparsevec \|_{1} = \frac{1}{\sqrt{2}}\| \sparseBlock \|_{F}^{(3 \times 2)}$.


\subsection{Network approximate solution} \label{sec: network_approx}
The optimization problem in Eq.~\eqref{eq: lower_level_problem} is an instance of dictionary learning problem with sparse codes $\sparsevec$. The classical approach to this problem is by solving the Iterative Shrinkage and Thresholding Algorithm (ISTA)~\cite{ista_1}. However, Papyan et al.~\cite{papyan2017convolutional} show that a single layer feedforward network with Rectified Linear Unit (ReLU) activations approximate one step of ISTA, with the bias terms $^{l}\boldsymbol{\lambda}$ adjusting the sparsity of recovered code for the $l^{\text{th}}$ layer. Furthermore, the dictionaries $[ \quad \hspace{-3.5mm} ^{1}\dict, \ldots,  \quad \hspace{-3.5mm} ^{L}\dict]$ can be learned by back-propagating through the feedforward network. We devise a network architecture that serves as an approximate solver to the above optimization problem and provide derivations in the following subsections.

\vspace{-0.4cm}
\paragraph{Approximating sparse codes.} We review the sparse dictionary learning problem and consider the single-layer case stated above. To reconstruct an input signal $\mathbf{X}$, the optimization problem becomes
\begin{align*}
    \min_{\sparseBlock} \|\mathbf{X} - \dict\sparseBlock\|_{F} + \boldsymbol{\lambda}\| \sparseBlock \|_{F}
\end{align*} As stated above, Papyan et al.~\cite{papyan2017convolutional} propose that one iteration of ISTA gives back the block-sparse codes $\sparseBlock$ as
\begin{align*}
    \sparseBlock = \text{ReLU} (\dict^{\top}\mathbf{X}; \boldsymbol{\lambda})
\end{align*} We interpret \text{ReLU} as solving for the block-sparse code and incorporate \text{ReLU} as the nonlinearity in our encoder part of the network. 

\vspace{-0.4cm}

\paragraph{Encoder architecture.} We propose to devise an encoder network $f_{e}$ that takes the 2D poses as input and outputs the block sparse codes $\sparseBlock$ that has within itself the rotation matrix $\cam$ as well as a rotationally invariant sparse code $\sparsevec$, \textit{i.e.} $f_{e}(\inputpose_{k}^{(n)}) \mapsto \Big( \quad \hspace{-3.5mm} ^{L}\sparseBlock_{k}^{*}\Big)$. Unrolling one iteration of ISTA for each layer, $f_{e}$ takes $\inputpose_{k}^{(n)}$ as 2D pose inputs and produces block sparse codes for the last layer $[ \quad \hspace{-3.5mm} ^{1}\sparseBlock_{k}^{(n)}, \ldots,  \quad \hspace{-3.5mm} ^{L}\sparseBlock_{k}^{(n)}]$ as output, shown in Fig.~\ref{fig:main_arch}


\begin{align}
\begin{split}
    ^{1}\sparseBlock_{k}^{{(n)}} &= \text{ReLU}\Big( \big[( \quad \hspace{-3.5mm} ^{1}\dict^{\#})^{\top} \cdot \inputpose_{k}^{(n)}\big]_{3 \times 2};  \quad \hspace{-3.5mm} ^{1}\boldsymbol{\lambda}^{{(n)}} \Big)  \\
    ^{2}\sparseBlock_{k}^{{(n)}} &= \text{ReLU}\Big(  (\quad \hspace{-3.5mm} ^{2}\dict \otimes \mathbf{I}_{3})^{\top} \cdot  \quad \hspace{-3.5mm} ^{1}\sparseBlock_{k}^{{(n)}}  ;  \quad \hspace{-3.5mm} ^{2}\boldsymbol{\lambda}^{{(n)}} \Big)  \\
    &\vdots  \\
    ^{L}\sparseBlock_{k}^{{(n)}} &= \text{ReLU}\Big(  \quad \hspace{-3.5mm} (^{L}\dict \otimes \mathbf{I}_{3})^{\top} \cdot \quad \hspace{-3.5mm} ^{L-1}\sparseBlock_{k}^{{(n)}}  ;  \quad \hspace{-3.5mm} ^{L}\boldsymbol{\lambda}^{{(n)}} \Big) \label{eq: encoder_part}    
\end{split}
\end{align} where $ \quad \hspace{-3.5mm} ^{l}\boldsymbol{\lambda}^{{(n)}}$ is the learnable threshold for each layer. $( \quad \hspace{-3.5mm} ^{l}\dict \otimes \mathbf{I}_{3})^{\top} \cdot \quad \hspace{-3.5mm} ^{l-1}\sparsevec_{k}^{{(n)}}$ is implemented by a convolution transpose.\vspace{-1.0em}

\paragraph{Rotation Factorization layer.}
At the bottleneck, our encoder network generates a block sparse code for $K-$views $ \quad \hspace{-3.5mm} ^{L}\sparseBlock_{k}^{(n)}$. As evident in Eq.~\eqref{eq: 2d_projection_equation}, since the block sparse code has rotations $\cam_{k}^{(n)}$ as well as an unrotated sparse code $\sparsevec_{k}^{(n)}$, we add a fully-connected layer that factorizes out these quantities, named Rotation Factorization (RF) layer, shown as a green block in Fig.~\ref{fig:main_arch}. Consequently, $^{L}\sparseBlock_{k}^{(n)}$ is then factorized into an unrotated sparse code $^{L}\sparsevec_{k}^{(n)}$ and the rotation matrix $\cam_{k}^{(n)}$ (constraining to $SO(3)$ using SVD) using this fully-connected RF layer. At this stage, we pool the features from all the rotationally invariant or unrotated sparse codes $^{L}\sparsevec_{k}^{(n)}$ using a sum pooling operation $\boldsymbol{g}$ that enforces the equivariance consistency within all the views by combining features from multiple views. 
\begin{align} \label{eq: pool}
 \boldsymbol{g}(^{L}\sparsevec_{1}^{{(n)}}, \ldots, ^{L}\sparsevec_{K}^{{(n)}}) \mapsto (^{L}\sparsevec^{{(n)}})   
\end{align}
as shown in architecture overview Fig.~\ref{fig:main_arch}, where $\boldsymbol{g}$ denotes a $\textbf{sum}$ operation. Since the pooled sparse code $^{L}\sparsevec$ is rotationally invariant, we generate a single canonical 3D structure $\struct$ through a decoder network $\boldsymbol{f}_{d}$, that remains equivariant to $K$ camera rotations $\cam_{1}, \ldots, \cam_{K}$. Thus, the decoder network $\boldsymbol{f}_{d}$ helps supervise the fully-connected RF layer.

\vspace{-0.3cm}
\paragraph{Insight behind multi-view consistency.}
For each individual view, we get a block sparse code representation $\sparseBlock_{k}^{(n)}$ that has the rotation $\cam_{k}^{(n)}$ combined with an unrotated sparse code $\sparsevec_{k}^{(n)}$. RF layer disentangles these quantities and generates codes that are consistent with an unrotated or canonicalized view. This architecture thus enforces equivariance consistency by consequently passing the unrotated sparse code $\sparsevec$ through a shape decoder to produce a canonicalized 3D structure. When we jointly encode multiple views into a single canonical shape, the equivariance is implicitly enforced after projecting them through the given multiple cameras. These multi-view projections help supervise the multi-view NRSfM network.

\vspace{-1.2em}

\paragraph{Decoder architecture.}

Finally, a decoder $\boldsymbol{f}_{d}$ is devised that takes input a pooled bottleneck sparse code (see Eq.~\ref{eq: pool}) and generates a canonical 3D structure $\struct$. Thus, $f_{d}( \quad \hspace{-3.5mm} ^{L}\sparsevec^{(n)}) \mapsto \big(\struct^{(n)}\big)$ 
\vspace{-0.25cm}
\begin{align}
    ^{L-1}\sparsevec^{(n)} &= \text{ReLU} ( \quad \hspace{-3.5mm} ^{L}\dict  \cdot \quad \hspace{-3.5mm} ^{L}\sparsevec^{(n)};  \quad \hspace{-3.5mm} ^{L}\boldsymbol{\lambda}^{(n)}) \nonumber \\
    & \vdots \nonumber \\
    ^{1}\sparsevec^{(n)} &= \text{ReLU} ( \quad \hspace{-3.5mm} ^{2}\dict \cdot \quad \hspace{-3.5mm} ^{2}\sparsevec^{(n)};  \quad \hspace{-3.5mm} ^{2}\boldsymbol{\lambda}^{(n)}) \nonumber \\
    \struct^{(n)} &=  \quad \hspace{-3.5mm} ^{1}\dict \cdot \quad \hspace{-3.5mm} ^{1}\sparsevec^{(n)} \label{eq: decoder_part}
    \vspace{-0.5em}
\end{align}
We analytically compute a closed-form solution to $\cam^{*}$ as a solution to an Orthographic-n-Point (OnP) problem that implicitly acts as supervisory signal for the $\cam_{k}^{(n)}$. Detailed proof is shown in the supplementary section. 

\vspace{-1.2em}
\paragraph{Calculating $\cam$ using solution from OnP} \label{sec: OnP_Para}
We are using a closed-form algebraic solution to $\cam^{*}$ that gives us an optimal solution for $\cam$ produced by the network at the bottleneck stage. We opt to use an algebraic solution that can be implemented as a
differentiable operator and could be easily accomplished via modern autograde packages. 
The detailed proof for the OnP solution is given in the appendix.

\vspace{-1.2em}
\paragraph{Loss function} To reemphasize the loss function in our neural architecture, the loss function driving the proposed approach is a reprojection error
\vspace{-1.0em}
\begin{align} \label{eq: final_loss}
    \mathcal{L} = \frac{1}{KN}\sum_{k=1}^{K}\sum_{n=1}^{N} \| \inputpose_{k}^{(n)} - \struct^{(n)}\cam_{k}^{(n)}  \|_{F}
\end{align}


\section{Experiments} \label{sec: results}
Evaluation over objects such as human body, hands, and monkey body is divided into two major categories: \textit{(i)} Multi-view 3D reconstruction of an input 2D dataset, and \textit{(ii)} Generation of 3D labels for unseen 2D data. The former compares against classical algorithms to generate high-fidelity 3D reconstruction from multi-view 2D input datasets. The latter discusses the generalization capability of our approach and shows that it does not overfit. For this, we follow one of the standard protocols for a human pose dataset and show results on the validation split. Equally competent 3D reconstructions are obtained for squishy deformable categories such as balloon deflation or paper tearing~\cite{jensen2021benchmark} -- making the proposed approach agnostic to deformable object categories. Results from NRSfM Challenge dataset~\cite{jensen2021benchmark} using 2 camera views are shown in the supplementary section.

\vspace{-0.5cm}
\paragraph{Network architecture and implementation details}
The same neural prior architecture is used for all the $K$ views across different datasets. We use $K-$encoders and a single shape decoder to generate one 3D structure in a canonicalized frame. The dictionary size (\textit{i.e.} neural units) within each layer of encoder is decreased exponentially: $\{1024,512,256,128,64,32,16,8\}$. Ideally, if a validation set with 3D groundtruth is provided, we could select optimal architecture based on cross-validation. However, due to the unsupervised setting, we rather set the hyperparameters heuristically. For the encoder and decoder architecture discussed in Eq.~\eqref{eq: encoder_part},~\eqref{eq: decoder_part}, we use a convolutional network as in Kong et al.~\cite{deep_nrsfm} and share the convolution kernels (\textit{i.e.} dictionaries) between the encoder and decoder. 

\vspace{-0.4cm}
\paragraph{Training details} We keep the same weightings for the reprojection error shown in loss function Eq.~\eqref{eq: final_loss}. We use the Adam optimizer~\cite{adamoptim} in our implementation.

\vspace{-0.4cm}
\paragraph{Evaluation metrics} Unless otherwise noted, we utilize the following metrics to assess the prediction accuracy of 3D reconstruction. \textbf{PA-MPJPE}: 
prior to computing the mean per-joint position error, we standardize the scale of the predictions by normalizing them to match against the given ground-truth (GT) followed by rigidly aligning these predictions to GT using Procrustes alignment. Lower the better. \textbf{PCK}: percentage of correct keypoints after Procrustes alignment. The predicted joint is viewed as correct if the separation between the predicted and the GT joint is within a specific range (usually in \textit{cm} or \textit{mm}). 

\vspace{-0.4cm}
\paragraph{Monkey body dataset}
\noindent OpenMonkeyStudio~\cite{openmonkey} is a huge Rhesus Macaque monkey pose dataset in a setup similar to PanOptic Studio where $62$ cameras capture the markerless pose of Rhesus Macaque monkeys. We use the provided 2D annotations over the Batch (7, 9, 9a, 9b, 10, and 11). This dataset also provides the groundtruth 3D labels for the given batches to evaluate the 3D reconstruction performance. 

\vspace{-0.4cm}
\paragraph{Human body dataset} Human 3.6 Million (H3.6M)~\cite{human36m} is a large-scale human pose dataset with images featuring actors performing daily activities from 4 camera views - annotated by motion capture systems. The 2D keypoint annotations of H3.6M preserve the perspective effect, and thus is a realistic dataset for evaluating the practical usage of generating 3D labels for unseen data as well as test the generalization capability of our approach. The results obtained on this dataset supports our hypothesis that the weak-perspective is a reasonable preliminary assumption; we plan to account for perspective effects as part of future work. We use this dataset for both quantitative and qualitative evaluation. For generating 3D reconstruction of an input dataset (see Sec.~\ref{sec: 5_1}), we pick 5 subjects (1, 5, 6, 7, 8) and compare against the classical multi-view triangulation baselines. For generating 3D labels over unseen 2D data to showcase the generalization capability (see Sec.~\ref{sec: generalization}), we follow the standard protocol on H3.6M and use the subjects (1, 5, 6, 7, 8) during the training stage and the subjects (9, 11) for evaluation stage. Evaluation is performed on every 64th frame of the test set. We include average errors for each method.

\vspace{-0.4cm}
\paragraph{Human hands dataset}
\noindent Finally, we use an open-source hands dataset - FreiHand~\cite{freihand} - a large-scale open-source dataset with varied movements of hands with 3D pose annotated by motion capture systems. It consists of $32560$ samples with their corresponding camera intrinsics. We generate random camera extrinsics and randomly create multiple camera views to generate multi-view 2D inputs for evaluating the proposed approach.

\subsection{3D reconstruction of an input 2D dataset} \label{sec: 5_1}
\noindent Like multi-view triangulation or bundle adjustment, our approach jointly infers the unknown 3D shape and camera rotations from 2D keypoints. By simultaneously fitting the shared network parameters used to recover shape and pose, our approach constrains the possible reconstructions much more strongly than multi-view triangulation or bundle-adjustment approaches. Although we showcase the generalization capability of the setup by applying the fitted network to generate 3D labels for unseen 2D data (see Sec.~\ref{sec: generalization}), the major contribution of our approach is the optimization process for multi-view 3D reconstruction of an input 2D dataset. The goal is to evaluate the robustness of the proposed multi-view neural shape prior across different shape variations and hence as part of the evaluation, we report how well our method is able to reconstruct different datasets compared to the baseline methods. 

\vspace{-0.4cm}
\paragraph{Baseline} 
We use an implementation of iterative multi-view triangulation with robust outlier rejection~\cite{hartley1997triangulation, hartley_multiview_geometry}, referred to as \textbf{TRNG} -- a method of choice for multi-view 3D human pose learning by Kocabas et al.~\cite{self-supervise-3d}. They also provide an open-source implementation for this baseline method. A more recent method doing classical optimization on triangulation is proposed by Lee and Civera~\cite{closed-form-trng}, however, their method is not necessarily optimal in terms of accuracy, but more in terms of computation time. \textbf{TRNG} first finds the points which minimize the distance from all the rays and removes the rays which are the furthest away from that point. It then re-evaluates the triangulation and this iteration is repeated 2-3 times. Empirically, we find that increasing the iteration leads us to predict near-perfect 3D reconstruction if we have exact camera calibration parameters and exact, clean 2D projections. We consider this to be a very strong baseline comparison since this approach is being widely used in industry as well as academia to generate accurate 3D reconstructions used to train 3D regression methods. We evaluate our approach on the above three datasets with substantial non-rigid deformities. For all the given experiments the cameras are chosen at random and the same set of cameras are used in the comparative baselines for a fair comparison. 
\begin{table*}[!t]
\begin{center}
\begin{tabular}{c | c c c c c c}
\hline
Method & Batch$\#7$ & Batch$\#9$ & Batch$\#9a$ & Batch$\#9b$ & Batch$\#10$ & Batch$\#11$ \\ 
\hline\hline
\textbf{TRNG}  & $21.21$ & $24.32$ & $30.67$ & $24.50$ & $26.10$ & $22.77$   \\
\textbf{MV NRSfM} & $\mathbf{8.36}$ & $\mathbf{8.25}$ & $\mathbf{9.12}$ & $\mathbf{11.52}$ & $\mathbf{8.203}$ & $\mathbf{8.17}$ \\
\hline
\end{tabular}
\end{center}
\vspace{-0.5cm}
\caption{\textbf{PA-MPJPE} error values for Monkey body dataset shows substantial improvement over the baseline multi-view triangulation approach while using only two views over noisy 2D keypoints. \textbf{PA-MPJPE} values are in \textbf{cm}.}
\label{tab: pa-mpjpe-monkey}
\end{table*}

\begin{table*}[h]
\begin{center}
\begin{tabular}{l|c|r|r|c|r|r|c|r|r}
\hline
\multirow{3}{*}{} & \multicolumn{9}{c}{\textbf{S1, S5, S6, S7, S8}}                                                                                                                                                                                  \\ \cline{2-10} 
                  & \multicolumn{3}{c|}{Extrinsics Noise}   & \multicolumn{3}{c|}{Intrinsics Noise}                                     & \multicolumn{3}{c}{2D keypoints Noise}                                \\ \cline{2-10} 
                  & $\sigma=0.1$                   & \multicolumn{1}{c|}{$\sigma=0.5$} & \multicolumn{1}{c|}{$\sigma=0.9$} & $\sigma=0.1$                    & \multicolumn{1}{c|}{$\sigma=0.5$} & \multicolumn{1}{c|}{$\sigma=0.9$} & $\sigma=15$                     & \multicolumn{1}{c|}{$\sigma=25$} & \multicolumn{1}{c}{$\sigma=35$} \\ \hline
{TRNG}          & $65.49$ &   $131.66$                       &      $145.94$                    & $69.57$ &            $188.63$             &           $234.47$              & $70.08$ &  $114.06$                      &         $154.41$               \\ \hline
2-Views (ours)  & \multicolumn{6}{c|}{$\mathbf{30.53}$} & $\mathbf{54.22}$ &              $\mathbf{65.74}$          &       $\mathbf{77.82}$             \\ \hline
\end{tabular}
\vspace{-0.8em}
\caption{Robustness to camera calibration and 2D annotations noise for Human 3.6M dataset. Values are in \textbf{mm}.}
\label{tab: human36m}
\end{center}
\end{table*}

\begin{table} 
\vspace{-1.5em}
\begin{center}
\begin{tabular}{c | c } 
\hline
\textbf{Method} & \textbf{PCK} \\ 
\hline\hline
2 Views~\cite{openmonkey} & $1.2 \%$  \\
4 Views~\cite{openmonkey} & $59 \%$ \\
8 Views~\cite{openmonkey} & $80 \%$ \\
16 Views~\cite{openmonkey} & $82 \%$ \\
32 Views~\cite{openmonkey} & $87 \%$ \\
48 Views~\cite{openmonkey} & $95 \%$ \\
2-views (ours) & $\mathbf{68.63} \%$ \\
3-views (ours) & $\mathbf{84.63} \%$ \\
\hline
\end{tabular} 
\end{center}
\vspace{-1.2em}
\caption{Percentage of Correct Keypoint (PCK) $\%$ for OpenMonkeyStudio dataset. Following~\cite{openmonkey}, the threshold for considering a keypoint to be correct is set at $10$ cm.}
\label{tab: pck_monkey}
\end{table}

\begin{figure}[!t]
	{\includegraphics[width=1\linewidth]{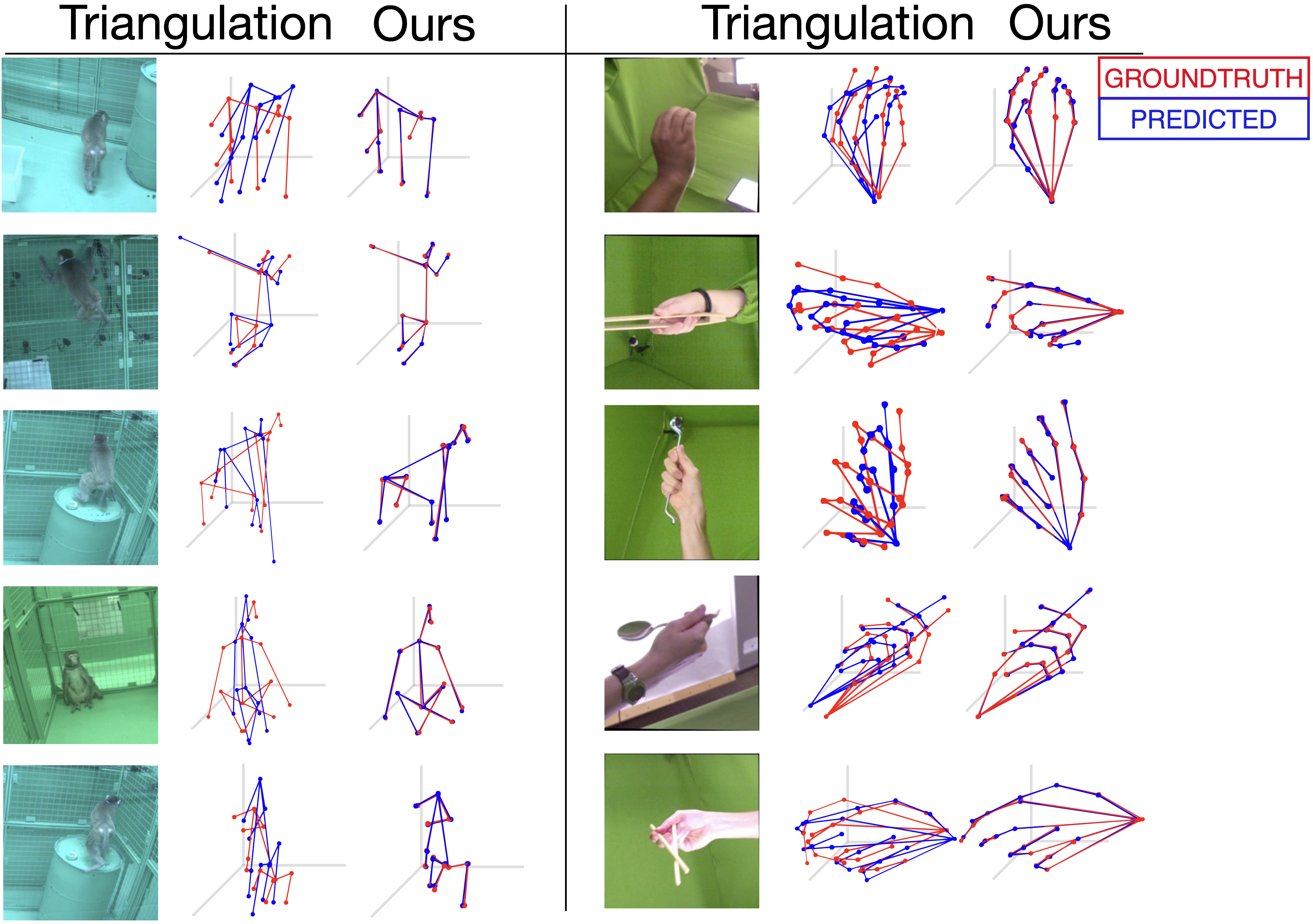}}
	\caption{Qualitative 3D reconstruction comparison between the multi-view triangulation technique and our technique for Monkey body~\cite{openmonkey} and human hands~\cite{freihand} when operated over noisy 2D keypoints.
	}
	\vspace{-0.4cm}
	\label{fig: monkey_hand_compare}
\end{figure}

\vspace{-0.4cm}

\paragraph{Evaluation analysis}

\begin{figure}[t]
	\centering
	\includegraphics[width=0.95\linewidth]{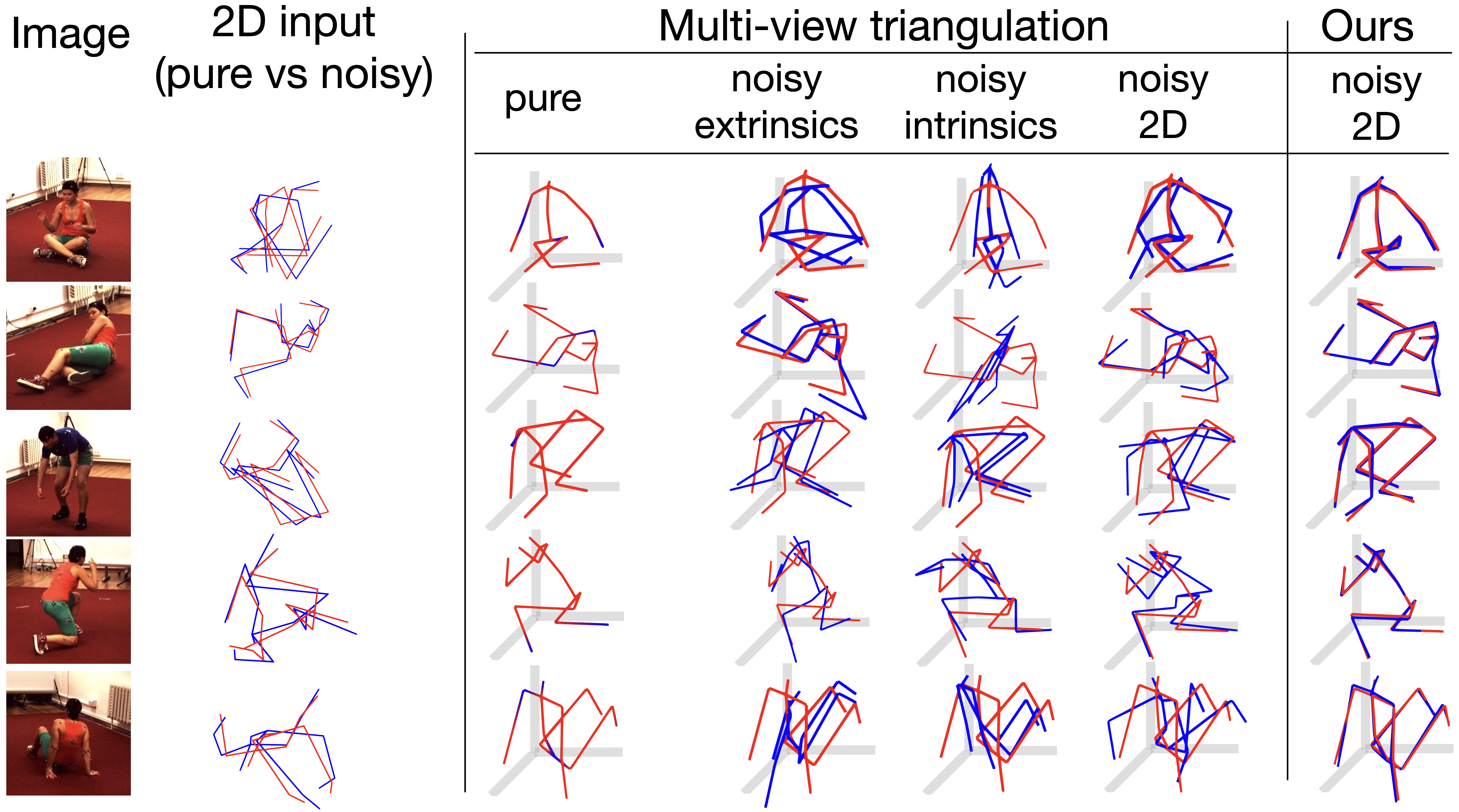}
	\caption{Qualitative results on Human 3.6M dataset with $\sigma=[0.5, 0.5, 25]$ as intrinsics, extrinsics, and 2D keypoints Gaussian noise, respectively.}
	\vspace{-0.4cm}
	\label{fig:human_36_compare}
\end{figure}

For the Monkey body dataset, multi-view 3D reconstruction with $2-$ or $3-$ view using our approach significantly outperforms the given results in~\cite{openmonkey} and achieves comparable fidelity with limited physical views. We consider all the keypoints as correct if their reconstruction is within $10 \textit{cm}$ of the groundtruth in the \textbf{PCK} protocol. Table~\ref{tab: pck_monkey} and top-right plot in Fig.~\ref{fig:overview_final} shows that we outperform the given results of 2-Views by a significant margin ($1.2\%$ vs. $68.63\%$). The fidelity of 3D reconstructions using our method continues to rise as we add in more views - evident by the uptick in performance from $3-$views. Qualitative performance of Monkey and Human hands dataset is shown in Fig.~\ref{fig: monkey_hand_compare} and quantitative performance of Monkey body is given in Tab.~\ref{tab: pa-mpjpe-monkey} when operated over noisy 2D keypoints. For the Human body dataset, we inject noise in the camera extrinsics, intrinsics, and 2D keypoints separately and compare the performance in Fig.~\ref{fig:human_36_compare} and Tab.~\ref{tab: human36m}. The baseline method fails when noise with a small standard deviation is added, degrading the fidelity of the 3D reconstruction. Since our approach is only dependent on the quality of 2D keypoints, it shows slightly degraded performance when the noise is injected over the input 2D keypoints. Qualitative 3D reconstruction of our approach in Fig.~\ref{fig: monkey_hand_compare},~\ref{fig:human_36_compare} shows the visual improvement over the classical multi-view triangulation approaches when operated over noisy 2D keypoints.

\subsection{Generation to unseen 2D data} \label{sec: generalization}
\vspace{-0.1cm}
\noindent Several multi-view approaches exists for 3D human pose estimation that leverage either full or weak 3D supervision~\cite{learnable-triangulation,remelli2020lightweight,kadkhodamohammadi2021generalizable,tome,Pavlakos,rhodin-monocular-from-multiview,self-supervise-3d}. None of these references, however, directly tackle the unsupervised multi-view 3D reconstruction problem and hence are not as general as our solution. However, to showcase the generalization capability of our approach, we include these approaches in our evaluation, shown in Tab.~\ref{tab: h36m_generalize} (the values are in \textbf{mm}). Furthermore, we also compare against recent monocular unsupervised 3D reconstruction methods. We leverage the processed datasets by Dovotny et al.~\cite{c3dpo} as the detected 2D keypoints for a fair evaluation and use the evaluation split of H3.6M dataset for this comparison. We find that our approach outperforms all other unsupervised approaches, and is on-par with many supervised methods.

\begin{table}[]
\vspace{-0.5cm}
\begin{center}
\begin{tabular}{lc|c}
\hline
\multicolumn{1}{l|}{\textbf{Method}}            & \multicolumn{1}{l|}{\textbf{Detected 2D}} & \multicolumn{1}{l}{\textbf{GT 2D} } \\ \hline
\rowcolor[HTML]{FFFFFF} 
\cellcolor[HTML]{FFCCC9}{Iskakov et al.~\cite{learnable-triangulation}} & 20.8                                      & -                                  \\
\rowcolor[HTML]{FFFFFF} 
\cellcolor[HTML]{FFCCC9}{Remelli et al.~\cite{remelli2020lightweight}} & 30.2                                      & -                                  \\
\rowcolor[HTML]{FFFFFF} 
\cellcolor[HTML]{FFCCC9}Kadkhodamohammadi et al.  ~\cite{kadkhodamohammadi2021generalizable}          & 49.1                                      & -                                  \\
\rowcolor[HTML]{FFFFFF} 
\cellcolor[HTML]{FFCCC9}Tome et al.    ~\cite{tome}         & 52.8                                      & -                                  \\
\rowcolor[HTML]{FFFFFF} 
\cellcolor[HTML]{FFCCC9}Pavlakos et al.   ~\cite{Pavlakos}      & 56.9                                      & -                                  \\
\rowcolor[HTML]{FFFFFF} 
\cellcolor[HTML]{FFCCC9}Multi-view Martinez ~\cite{martinez}    & 57.0                                      & -                                  \\
\rowcolor[HTML]{FFFFFF} 
\cellcolor[HTML]{FFCCC9}{Rhodin et al.}~\cite{rhodin-monocular-from-multiview}             & {\color[HTML]{333333} 51.6}  & {\color[HTML]{333333} -}     \\
\rowcolor[HTML]{FFFFFF} 
\cellcolor[HTML]{FFCCC9}{Kocabas et al ~\cite{self-supervise-3d}}       & {\color[HTML]{333333} 45.04} & {\color[HTML]{333333} -}     \\
\rowcolor[HTML]{FFFFFF} 
\cellcolor[HTML]{FFF600}{Kocabas et al. (SS w/o R) ~\cite{self-supervise-3d} } & {\color[HTML]{333333} 70.67} & {\color[HTML]{333333} -}     \\ 
\rowcolor[HTML]{FFFFFF} 
\cellcolor[HTML]{FFF600}{PRN}                 ~\cite{PRN}                             & 124.5                        & 86.4                         \\
\rowcolor[HTML]{FFFFFF} 
\cellcolor[HTML]{FFF600}{RepNet}           ~\cite{repnet}                                & 65.1                         & 38.2                         \\
\rowcolor[HTML]{FFFFFF} 
\cellcolor[HTML]{FFF600}{Iqbal et al.~\cite{jan_kautz_mv_weak_supervised}}                                     & 69.1                         & -                            \\ 
\rowcolor[HTML]{FFFFFF}
\rowcolor[HTML]{FFFFFF} \rowcolor[HTML]{FFFFFF} 
\cellcolor[HTML]{FFF600}{Pose-GAN ~\cite{pose-gan} }       & 173.2                                     & 130.9                              \\
\rowcolor[HTML]{FFFFFF} 
\cellcolor[HTML]{FFF600}Deep NRSfM  ~\cite{deep_nrsfm}          & -                                         & 104.2                              \\
\rowcolor[HTML]{FFFFFF} 
\cellcolor[HTML]{FFF600}C3DPO  ~\cite{c3dpo}                 & 153.0                                     & 95.6                               \\ \hline
\multicolumn{1}{l|}{\cellcolor[HTML]{FFF600}MV NRSfM (Ours)} & \multicolumn{1}{c|}{\textbf{45.2}} &    \textbf{30.2}   \\ \hline
\end{tabular} 
\end{center}
\vspace{-0.4cm}
\caption{Generalization experiments. Red tint rows have 3D supervision. Yellow tint are unsupervised 3D reconstruction methods. Our method is on par with most 3D supervised methods, and outperforms all unsupervised methods.}
\vspace{-0.25cm}
\label{tab: h36m_generalize}
\end{table}

\section{Discussion and Conclusion} \label{sec: conclusion}

\vspace{-0.18cm}
We propose a multi-view 2D-3D lifting architecture that incorporates neural shape prior using the recent advances of modern deep learning methods. Our contribution combines the ideas from multi-view geometry and recent monocular deep 3D lifting approaches -- essentially leveraging the best features of both worlds. We also show the generalization capability of the proposed approach by generating accurate 3D reconstructions on unseen data. Although we require limited rigid views at any instant of time, our approach still requires multiple non-rigid atemporal views to enforce the proposed neural shape prior during training/optimization. Literature in domain of neural shape priors is extensive~\cite{c3dpo,deep_nrsfm_pp} and new innovations are proposed constantly, and we believe we could leverage these innovations in our framework as part of future direction.

\newpage

{\small
\bibliographystyle{ieee_fullname}
\bibliography{main}
}

\end{document}